\documentclass[letterpaper, 10 pt, conference]{ieeeconf}  

\IEEEoverridecommandlockouts
\usepackage{graphics} 
\usepackage{epsfig} 
\usepackage{mathptmx} 
\usepackage{times} 
\usepackage{amsmath} 
\usepackage{amssymb}  
\usepackage[utf8]{inputenc} 

\usepackage{diagbox}
\usepackage{multicol}
\usepackage{threeparttable}
\usepackage{multirow}
\usepackage{color}
\usepackage{array}
\usepackage{setspace}
\usepackage{makecell}
\usepackage{subcaption}
\usepackage[pagebackref,breaklinks,colorlinks]{hyperref}

\title{\LARGE \bf
JSTR: Joint Spatio-Temporal Reasoning for \\ Event-based Moving Object Detection
}

\author{Hanyu Zhou, Zhiwei Shi, Hao Dong, Shihan Peng, Yi Chang, and Luxin Yan$^{*}$
\thanks{$^{*}$Corresponding author.}
\thanks{Hanyu Zhou, Zhiwei Shi, Hao Dong, Shihan Peng, Yi Chang and Luxin Yan are with National Key Lab of Multispectral Information Intelligent Processing Technology, School of Artificial Intelligence and Automation, Huazhong University of Science and Technology, Wuhan, China. Email: \{hyzhou, shizhiwei, donghao0205, pengshihan, yichang, yanluxin\}@hust.edu.cn
}}

\begin{document}

\maketitle
\thispagestyle{empty}
\pagestyle{empty}

\begin{abstract}
Event-based moving object detection is a challenging task, where static background and moving object are mixed together. Typically, existing methods mainly align the background events to the same spatial coordinate system via motion compensation to distinguish the moving object. However, they neglect the potential spatial tailing effect of moving object events caused by excessive motion, which may affect the structure integrity of the extracted moving object. We discover that the moving object has a complete columnar structure in the point cloud composed of motion-compensated events along the timestamp. Motivated by this, we propose a novel joint spatio-temporal reasoning method for event-based moving object detection. Specifically, we first compensate the motion of background events using inertial measurement unit. In spatial reasoning stage, we project the compensated events into the same image coordinate, discretize the timestamp of events to obtain a time image that can reflect the motion confidence, and further segment the moving object through adaptive threshold on the time image. In temporal reasoning stage, we construct the events into a point cloud along timestamp, and use RANSAC algorithm to extract the columnar shape in the cloud for peeling off the background. Finally, we fuse the results from the two reasoning stages to extract the final moving object region. This joint spatio-temporal reasoning framework can effectively detect the moving object from motion confidence and geometric structure. Moreover, we conduct extensive experiments on various datasets to verify that the proposed method can improve the moving object detection accuracy by 13\%.

\end{abstract}

\begin{figure}
  \setlength{\abovecaptionskip}{5pt}
  \setlength{\belowcaptionskip}{-5pt}
  \centering
   \includegraphics[width=1.0\linewidth]{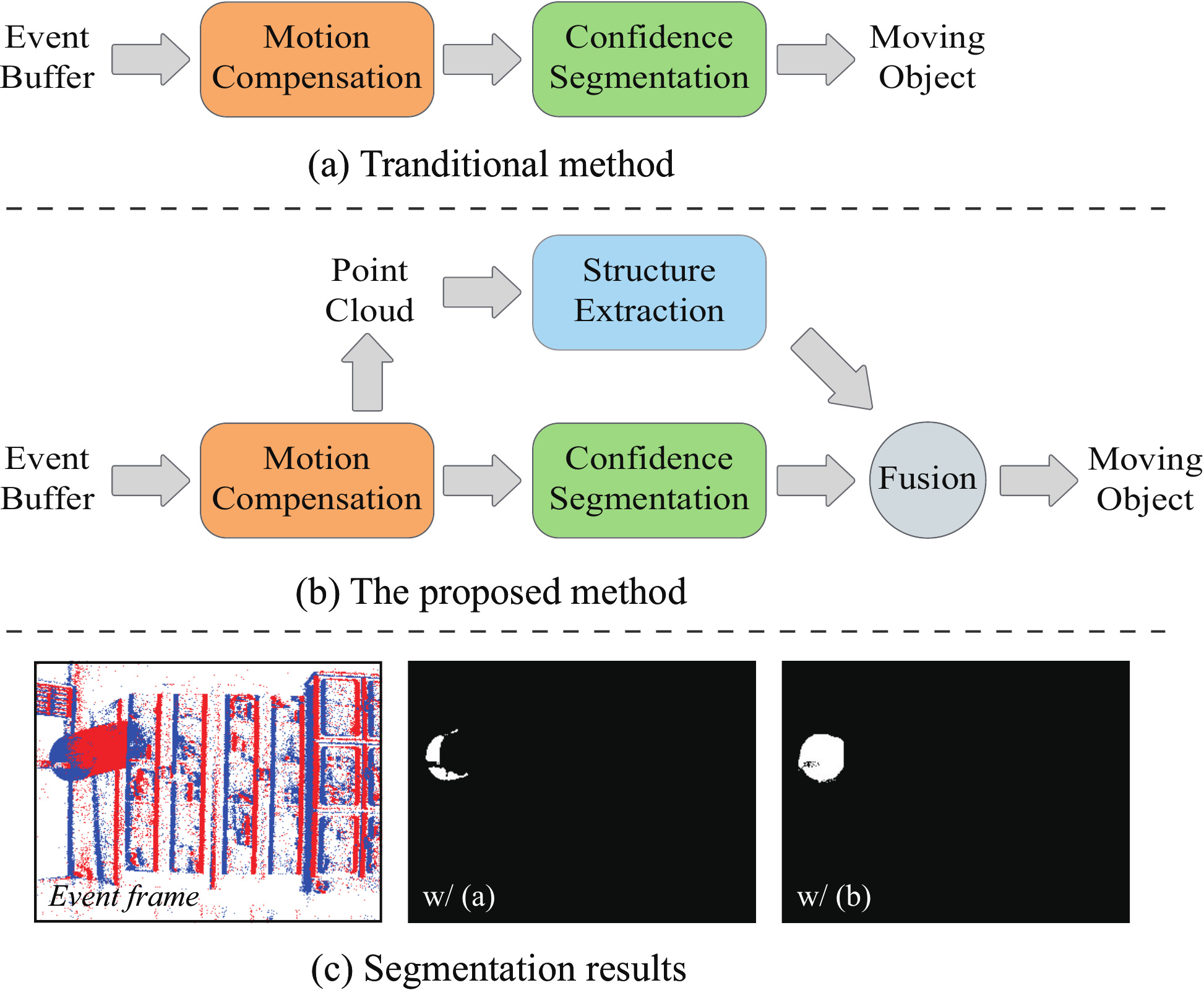}
   \caption{Illustration of different moving object detection frameworks. (a) Traditional method. (b) The proposed joint spatio-temporal reasoning framework. (c) Corresponding segmentation results. Traditional method reasons the motion confidence distribution from spatial dimension for moving object, while suffering from the incomplete contour. Joint spatio-temporal reasoning can further exploit the structure characteristic of moving object from the temporal event point cloud for structure integrity of moving object.
   }
   \label{Paradim}

\end{figure}

\section{INTRODUCTION}
Event camera \cite{gallego2020event} is a neuromorphic vision sensor with the advantage of low latency and high dynamic range, asynchronously triggers events due to illumination change, applied in space situational awareness \cite{jawaid2023towards, cohen2019event, afshar2020event} and robot obstacle avoidance \cite{falanga2020dynamic}.
Event-based moving object detection is a challenging task. When static camera platform, event camera only senses moving objects. When dynamic camera platform, it is difficult to detect moving objects since the background events generated by ego motion are mixed with the moving object events.

Existing methods mainly take optimization strategy \cite{parameshwara2021spikems, stoffregen2019event, mitrokhin2018event, lu2021event, sanket2019evdodge} or inertial measurement unit (IMU) \cite{falanga2020dynamic, forrai2023event, zhao2023event} to align background events with different timestamps to the image spatial coordinate at the same timestamp via motion compensation to distinguish moving objects.
For example, Mitrokhin \emph{et al.} \cite{mitrokhin2018event} formulated 3D geometric relationships within event stream as an optimization model, and solved the ego motion parameters via iterative optimization for aligning background events. Sanket \emph{et al.} \cite{sanket2019evdodge} fitted the homography transformation matrix between adjacent event frames using CNN, and warped the current event frame to the reference for distinguishing motion.
These optimization-based methods rely heavily on high-quality event data, while suffering degradation when applied in adverse conditions (\emph{e.g.}, nighttime scene). Falanga \emph{et al.} \cite{falanga2020dynamic} and Zhao \emph{et al.} \cite{zhao2023event} utilized IMU to directly compensate background motion, avoiding the problem of degraded events. However, in Fig. \ref{Paradim} (a), these moving object detection methods neglect the potential spatial tailing effect of the moving object due to excessive motion, which affects the structure integrity of the extracted moving object in Fig. \ref{Paradim} (c). In this work, our goal (seeing Fig. \ref{Paradim} (b)) is to further improve the structure of the extracted moving object from event point cloud in Fig. \ref{Paradim} (c).

To address this issue, we map the motion-compensated events into the event frame in Fig. \ref{Analysis} (a), and transform the motion-compensated events as the point cloud along the timestamps in Fig. \ref{Analysis} (b). We observe that, in the spatial event frame, the background is sharp while the moving object has a serious tailing effect; in the temporal point cloud, the background is globally discrete while the moving object shows a complete columnar structure. Therefore, \emph{we suggest that extracting the regular structure from event point cloud benefits to promoting event-based moving object detection}.

In this work, we propose a joint spatio-temporal reasoning method for event-based moving object detection, as shown in Fig. \ref{Framework}. Specifically, we first calculate the rotation matrix from the IMU via Rodrigues algorithm \cite{mebius2007derivation}, and then use the rotation matrix to project events with different timestamps to the current timestamp through motion compensation. In the spatial reasoning stage, we map the compensated events to the same image coordinate, discretize the event timestamps to generate a motion confidence map (namely time image), and then segment the time image for the moving object via adaptive threshold. In the temporal reasoning stage, we transform the events into a point cloud along the timestamp, and take the RANSAC algorithm to extract the regular columnar structure in the point cloud for the moving object. Finally, we propose a contour-based fusion module to fuse the results of the spatial and temporal reasoning strategies to obtain the final moving object region. This joint spatio-temporal reasoning method can effectively detect the moving object from motion confidence and geometry structure. Moreover, we propose an event dataset to verify the effectiveness of the proposed method under various illumination conditions. Overall, the main contributions are as follows:
\begin{itemize}

\item We propose a joint spatio-temporal reasoning framework (JSTR) for event-based moving object detection, which can effectively extract the moving object with the complete regular structure.

\item We discover that the moving object has a significant columnar structure in the event point cloud, which motivates us to design a RANSAC-based cloud structure extraction module for the moving object.

\item We propose an event dataset for moving object detection, and verify the superiority of the proposed method on both the proposed and public event datasets.

\end{itemize}

\section{RELATED WORK}
\subsection{Event-based Moving object Detection}
Event moving object detection aims to extract the independent moving object from dynamic event stream. An intuitive method is object detection \cite{parameshwara2021spikems, nagaraj2023dotie}, which used the ground truth to train a deep model for object detection. However, this method cannot distinguish whether the motion of the object is caused by the object itself or the ego motion of the camera. Another method is motion compensation-based moving object detection \cite{parameshwara2021spikems, stoffregen2019event, mitrokhin2018event, lu2021event, sanket2019evdodge, falanga2020dynamic, forrai2023event, zhao2023event}, which mainly aligned the background events into the same image spatial coordinate to distinguishing the moving object using the motion obtained by optimization or assistant hardware.
For example, Sanket \emph{et al.} \cite{sanket2019evdodge} utilized a deep model to fit the homography matrix. Mitrokhin \emph{et al.} \cite{mitrokhin2018event} constructed a motion energy function to iteratively optimize motion parameters. Stoffregen \emph{et al.} \cite{stoffregen2019event} designed a local motion clustering algorithm to obtain the motion of different moving objects. These optimization-based methods rely heavily on high-quality event data, but may suffer degradation when applied in adverse conditions (\emph{e.g.}, nighttime scenes). Recently, IMU-based motion compensation methods have received attention. Forrai \emph{et al.} \cite{forrai2023event} and Zhao \emph{et al.} \cite{zhao2023event} calculated the rotation matrix from IMU via Rodrigues algorithm for compensating the background motion, relieving the burden on high-quality event data. However, these motion compensation-based moving object detection approaches neglect the potential spatial tailing effect of moving object events, limiting the structure integrity of the extracted moving object. To overcome this, we further exploit the structure characteristic of the moving object in the event cloud to improve moving object detection.

\begin{figure}
  \setlength{\abovecaptionskip}{5pt}
  \setlength{\belowcaptionskip}{-5pt}
  \centering
   \includegraphics[width=1.0\linewidth]{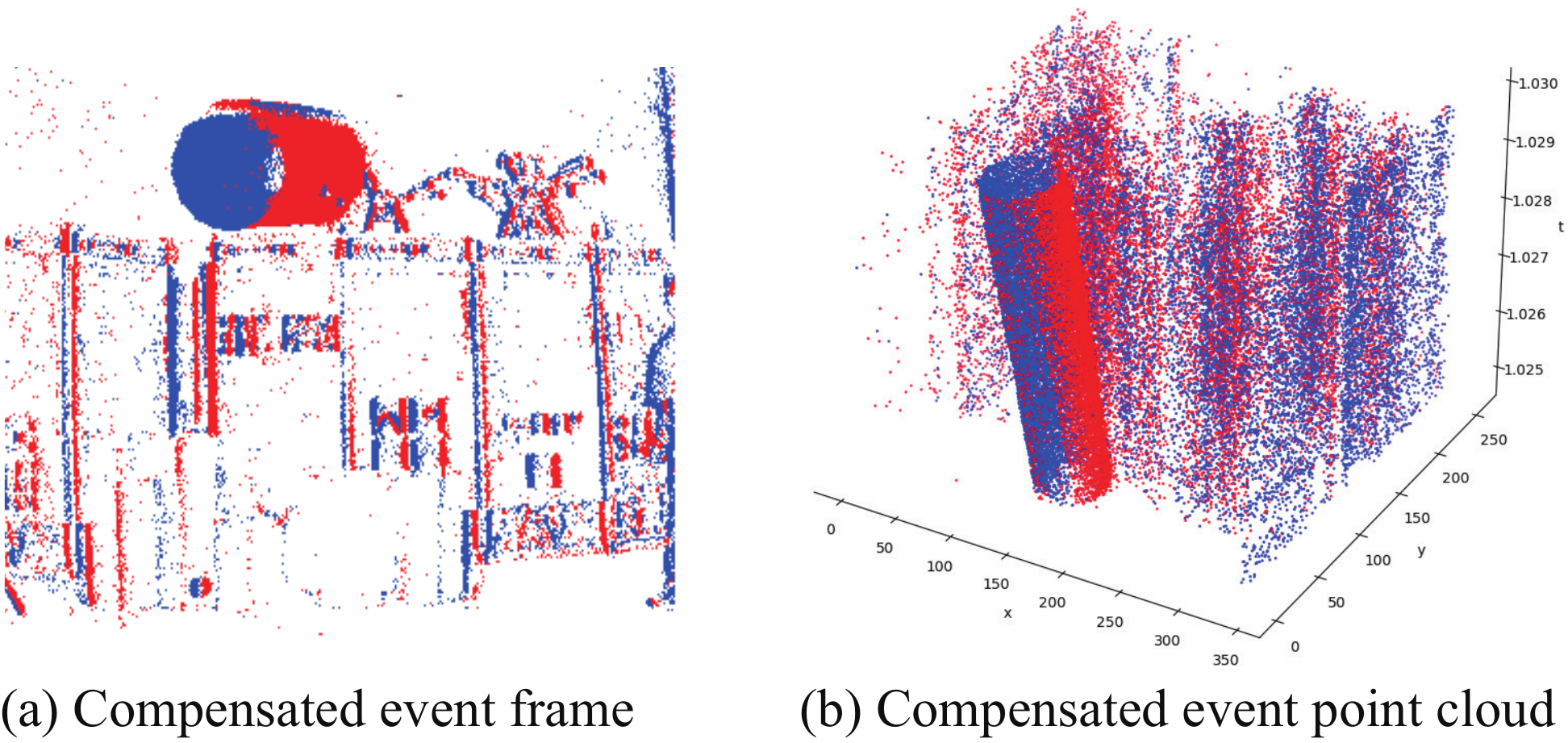}
   \caption{Visualization of event frame and point cloud. (a) Compensated event frame. (b) Compensated event point cloud. Moving object has an obvious tailing effect in the event frame, while has a significant columnar structure in the event cloud. This motivates us to improve moving object detection from the structure integrity of moving object.
   }
   \label{Analysis}
 \end{figure}

\begin{figure*}
  \setlength{\abovecaptionskip}{5pt}
  \setlength{\belowcaptionskip}{-5pt}
  \centering
   \includegraphics[width=0.98\linewidth]{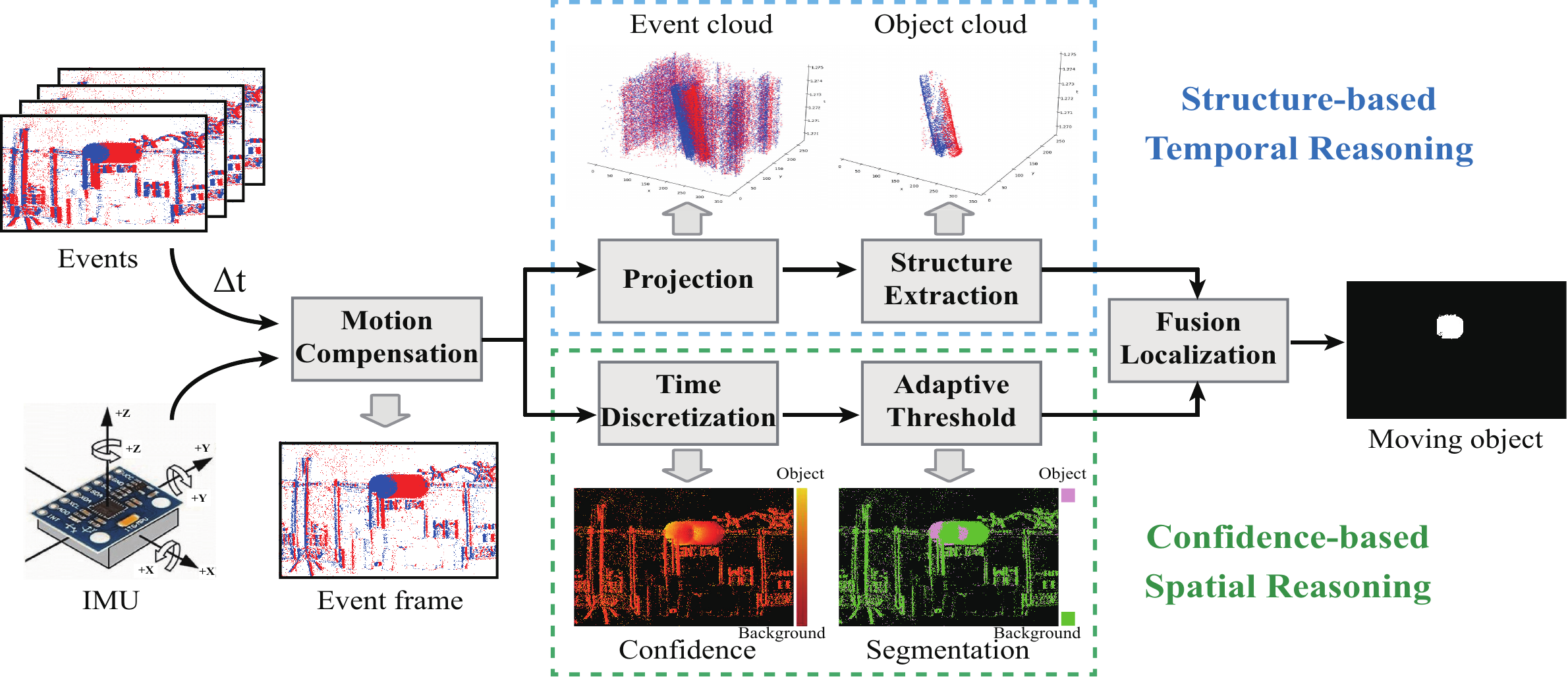}
   \caption{The framework of joint spatio-temporal reasoning (JSTR) mainly contains confidence-based spatial reasoning and structure-based temporal reasoning. Motion compensation aligns the background events. Confidence-based spatial reasoning segments the moving object region via adaptive threshold segmentation on time image (namely motion confidence map). Structure-based temporal reasoning further extracts the complete contour of the moving object from the event cloud.
   }
   \label{Framework}
 \end{figure*}

\subsection{Point Cloud Regular Structure Extraction}
The moving object has a columnar structure in the event point cloud, while the background is globally discrete. Our goal is to fit the regular structure and peel off the outliers in the event point cloud. A novel method is RANSAC \cite{fischler1981random}, which randomly samples some points from the data set and iteratively optimizes the parameters of the pre-defined objective function of the regular structure to obtain the optimal model, whose output is the points from the regular structure. Tao \emph{et al.} \cite{tao2023distinctive} designed a registration model to map the correspondence between different point clouds via RANSAC. Liu \emph{et al.} \cite{liu2013cylinder} used RANSAC to fit the circle shape. Zhang \emph{et al.} \cite{zhang2023high} removed the mismatched points during video stabilization using RANSAC. In this work, we are the first to introduce RANSAC to extract the regular structure of the moving object in the event point cloud.

\section{EVENTS DIFFERENCE ANALYSIS}
Event-based moving object detection is to distinguish the moving object from the background, and the key is to model their difference. We map the compensated events to the same spatial image coordinate, and transform the compensated events to the temporal point cloud coordinate along the timestamp. As shown in Fig. \ref{Analysis} (a), motion-compensated background events show a clear boundary, while there exists an obvious tailing effect of the moving object events. The main reason is that, motion compensation is to align the background events caused by ego motion, not the independent moving object. Event camera with high temporal resolution can trigger a large number of moving object events, resulting in the tailing effect that affects the structure integrity of the extracted moving object. As shown in Fig. \ref{Analysis} (b), in the event point cloud, the background events are globally discrete, while the moving object has a significant 3D columnar structure. This is because events of  moving object can still maintain their inertial motion in a short period, resulting in regular structure in the point cloud. Extracting such regular shape is beneficial to improving the structure integrity of the extracted moving object. Therefore, in Fig. \ref{Framework}, we predict the position of the moving object from the spatial event frame, and fine-tune the structure of the moving object from the temporal event point cloud.

\section{JOINT SPATIO-TEMPORAL REASONING}
As shown in Fig. \ref{Framework}, we input a dynamic event stream and align the background events via motion compensation. Then, we segment the region of the moving object from the compensated events with the confidence-based spatial reasoning, and refine the structure of the moving object from the compensated events with the structure-based temporal reasoning. In this section, we will introduce the necessary data association and the specific module construction.

\subsection{IMU-based Motion Compensation}
The essence of motion compensation is to calculate the pose transformation between adjacent timestamps (target and reference), and warp the pixel coordinate of the target timestamp to the reference timestamp via the geometry projection relationship \cite{zhou2017unsupervised}. Given a dynamic event stream $C$ in a certain period, we can obtain the event $e_t=[\textbf{x}, t, p]$ and the corresponding angular velocity $w_t$ from IMU, where $\textbf{x}=\{x, y\}$, $t$, $p$ are the pixel coordinate, timestamp, and polarity of the event, respectively.
We first calculate the average angular velocity $\overline{w}=\sum_{\delta{t}}w_t$, and then calculate the three-axis Euler angles $\alpha$, $\beta$, $\gamma$ of the event $e_t$ relative to the reference timestamp $t_0$, where $\alpha=\overline{w}_x(t-t_0)$, $\beta=\overline{w}_y(t-t_0)$, $\gamma=\overline{w}_z(t-t_0)$. We obtain the rotation matrix $R$ of the event $e_t$ relative to the reference timestamp $t_0$ via Rodrigues equation:
\begin{equation}
  \setlength\abovedisplayskip{3pt}
  \setlength\belowdisplayskip{3pt}
\begin{aligned}
R = R_z(\gamma) \cdot R_y(\beta) \cdot R_x(\alpha),
 \label{eq:rodrigues}
 \end{aligned}
\end{equation}
where $R_x$, $R_y$ and $R_z$ denote the rotation matrix with the Euler angles $\alpha$, $\beta$ and $gamma$, respectively. Then, we transform the pixel coordinate $\textbf{x}$ of the event $e_t$ to the reference timestamp via the geometry projection:
\begin{equation}
  \setlength\abovedisplayskip{3pt}
  \setlength\belowdisplayskip{3pt}
\begin{aligned}
 \hat{\textbf{x}} = KRK^{-1}\textbf{x},
 \label{eq:projection}
 \end{aligned}
\end{equation}
where $K$ is camera internal parameter. Note that, camera translation is the same as the above transformation process, while the input is acceleration value rather than angular velocity. Thus, the compensated event stream is as follows:
\begin{equation}
  \setlength\abovedisplayskip{3pt}
  \setlength\belowdisplayskip{3pt}
\begin{aligned}
 \hat{C} = \{\hat{x}, \hat{y}, t_0\} \forall \{x, y, t\} \in C.
 \label{eq:compensated_event}
 \end{aligned}
\end{equation}

\subsection{Confidence-based Spatial Reasoning}
In the compensated event stream, object motion includes ego motion and independent motion, and there exists an obvious event density difference between the moving object and the background. That is to say, the higher the event density,  the higher the moving object confidence. This motivates us to design a confidence-based spatial reasoning module.

\noindent \textbf{Time Discretization.} To model the event density difference between the moving object and the background, we follow Falanga \emph{et al.} \cite{falanga2020dynamic} to discretize the timestamps of the events. We first count the number of compensated events at different pixels from target timestamp $t$ to reference timestamp $t_0$:
\begin{equation}
  \setlength\abovedisplayskip{3pt}
  \setlength\belowdisplayskip{3pt}
\begin{aligned}
 \Omega_{ij} = \{\{\hat{x}, \hat{y}, t\}:\{\hat{x}, y, t_0\} \in \hat{C}, i = \hat{x}, j = \hat{y}\},
 \label{eq:count_image}
 \end{aligned}
\end{equation}
where $i$, $j$ denote the x,y-axis pixel coordinates of the events. Then, we construct the count image $I_{ij}=|\Omega_{ij}|$, and calculate the time image as follows:
\begin{equation}
  \setlength\abovedisplayskip{3pt}
  \setlength\belowdisplayskip{3pt}
\begin{aligned}
  T_{ij} = \frac{1}{I_{ij}} \sum\nolimits t : t \in \Omega_{ij}.
 \label{eq:time_image}
 \end{aligned}
\end{equation}

Since the pixel value of the time image will become larger as the larger timestamp, the subsequent image process will be blocked. To this end, we normalize the time image to $[-1, 1]$ for motion confidence as follows:
\begin{equation}
  \setlength\abovedisplayskip{3pt}
  \setlength\belowdisplayskip{3pt}
\begin{aligned}
  \rho (i, j) = T(t:t \in \Omega_{ij}) - \phi (T) / \delta t,
 \label{eq:motion_confidence}
 \end{aligned}
\end{equation}
where $\phi (T)$ denotes the mean operation.

\noindent \textbf{Adaptive Threshold Segmentation.} The normalized motion confidence map can reflect the probability of the moving object, and the key is how to reasonably set a threshold to segment the moving object. An intuitive strategy is to set a fixed threshold for segmentation. However, this strategy is not efficient. The main reason is that various ego motion can change the probability of the background, disturbing the impact of the segmentation threshold on the moving object extraction. Therefore, we introduce a linear function to adaptively adjust the segmentation threshold according to the IMU data as follows:
\begin{equation}
  \setlength\abovedisplayskip{3pt}
  \setlength\belowdisplayskip{3pt}
\begin{aligned}
  \tau (\omega) = a \cdot ||\omega|| + b,
 \label{eq:threshold}
 \end{aligned}
\end{equation}
where $\omega$ denotes the square root of three-axis angular velocity, and $a$, $b$
are the pre-defined parameters. $a$ is a weight that adapts to the angular velocity, and $b$ adjusts the classification threshold on moving object and background.

\noindent \textbf{Morphological Filter.}
Adaptive threshold can allow the initial segmentation result, while there may exist the potential background events mixed with the moving object. To further remove the background events, we introduce a contour-based morphological filter strategy, which models the spatial shape difference between the moving object and the residual background. Specifically, we first use Sobel operator to extract the contours of the segmented image. Then, we construct a fixed-size sliding window with all one values and use the window to traverse all contours, where the inner product is utilized to measure the pixel distribution discrepancy between each contour and the sliding window. The smaller the distribution discrepancy, the more likely the moving object and should be retained; conversely, the background should be filtered. After contour-based morphological filter, the segmentation image basically retains the independent moving object regions.

\subsection{Structure-based Temporal Reasoning}
Confidence-based spatial reasoning could promise a preliminary segmentation result of the moving object, while the structure integrity of the extracted moving object is limited by the object tailing effect. Combined with our observation in Fig. \ref{Analysis}, we propose a structure-based temporal reasoning in the event point cloud. Next, we will describe the columnar structure extraction method in detail.

\noindent \textbf{RANSAC-based Columna Structure Extraction.}
Moving object has a complete geometry structure in the event point cloud. Typical methods for extracting regular structure include Hough transformation \cite{duda1972use}, RANSAC \cite{fischler1981random}, etc. The essence of RANSAC lies in random sampling consensus and iterative optimization, which maximizes the extraction of points covered by the moving object structure in the event point cloud. Given a compensated event stream, we transform the events to the point cloud $D$ along the timestamp. We fit the columnar structure in the event point cloud as follows:

\emph{Step 1:} We first randomly sample a set 3D points $P$ from the event point cloud $D$. Since the moving object is a regular columnar structure, we set the points $P$ as $\{p_1, p_2, p_3, p_4\}$.

\emph{Step 2:} Columnar structure could be modeled by axis and radius characterization. We compute the normal vector of the columnar structure as its axis $\vec{n_z}$:
\begin{equation}
  \setlength\abovedisplayskip{3pt}
\begin{aligned}
  \vec{n_1} = \frac{p_2 - p_1}{||p_2-p_1||_2}, \vec{n_2} = \frac{p_3 - p_1}{||p_3-p_1||_2},
 \label{eq:normal_vector_1}
 \end{aligned}
\end{equation}
\begin{equation}
  \setlength\belowdisplayskip{3pt}
\begin{aligned}
  \vec{n_z} = \vec{n_1} \times \vec{n_2}.
 \label{eq:normal_vector}
 \end{aligned}
\end{equation}

\emph{Step 3:} We calculate the distance between the point $p_4$ and the normal vector $\vec{n_z}$ as the radius of the columnar structure:
\begin{equation}
  \setlength\abovedisplayskip{3pt}
  \setlength\belowdisplayskip{3pt}
\begin{aligned}
  \vec{n_3} = \frac{p_4 - p_1}{||p_4 - p_1||_2}, r_z = \frac{||\vec{n_z} \times \vec{n_3}||_2}{||\vec{n_3}||_2}.
 \label{eq:radius}
 \end{aligned}
\end{equation}

\emph{Step 4:} We further compute the distance between the other point $p_i$ and the constructed model:
\begin{equation}
  \setlength\abovedisplayskip{3pt}
  \setlength\belowdisplayskip{3pt}
\begin{aligned}
  \vec{n_i} = \frac{p_i - p_4}{||p_i - p_4||_2}, r(p_i) = \frac{||\vec{n_z} \times \vec{n_i}||_2}{||\vec{n_i}||_2}.
 \label{eq:model}
 \end{aligned}
\end{equation}

\emph{Step 5:} During one iteration, the whole event point cloud data is fed into the objective function that can determine the number of inliers:
\begin{equation}
  \setlength\abovedisplayskip{3pt}
  \setlength\belowdisplayskip{3pt}
\begin{aligned}
  max |\{p_i | |r(p_i)-r_z| \leq \theta, p_i \in D\}|.
 \label{eq:object_fucntion}
 \end{aligned}
\end{equation}
where $\theta$ is the distance threshold. We iteratively optimize the object function Eq. \ref{eq:object_fucntion} from \emph{step 1} to \emph{step 5} to obtain the optimal number of inliers, and the corresponding inliers are the event points from the moving object structure. Thus, RANSAC could fit the complete regular structure of the moving object, and benefit to relieve the limitation of the confidence-based spatial reasoning alone.

\begin{figure*}
  \setlength{\abovecaptionskip}{5pt}
  \setlength{\belowcaptionskip}{-5pt}
  \centering
   \includegraphics[width=0.99\linewidth]{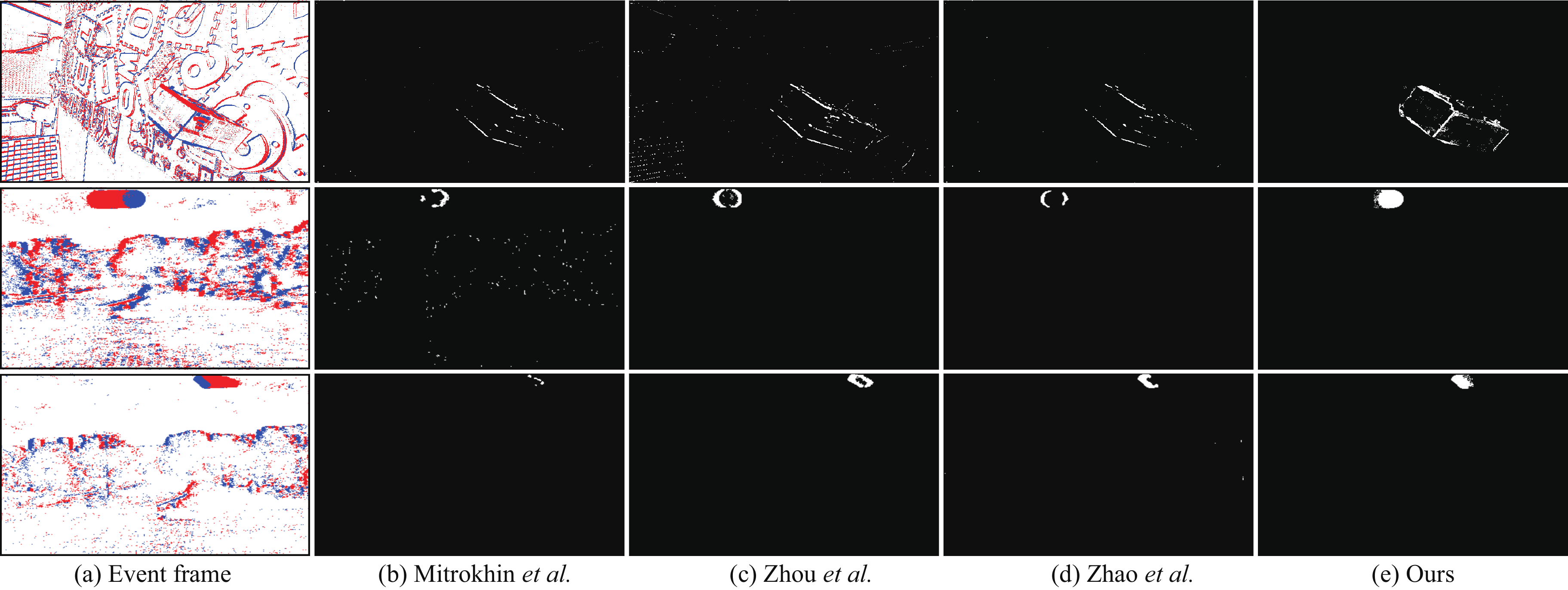}
   \caption{Visual comparison of moving object detection on public (first row) and self-collected (last two row) datasets.
   }
   \label{Comparison}
 \end{figure*}

\noindent \textbf{Moving Object Fusion Localization.}
The spatial reasoning can provide the position of the independent moving object, and the key is how to use the result of the temporal reasoning to ensure the structure integrity of the moving object. In this part, we design a contour-based fusion localization strategy. First, we back-project the extracted moving object point cloud into the image coordinate to obtain a clear contour. Next, we take this contour as the maximum boundary of the connected component, and expand the corresponding position of the moving object segmented by the spatial reasoning to the surrounding via connected component searching strategy \cite{he2009fast}. If the boundary is exceeded during searching process, we take the region within the boundary of the connected component as the final moving object region. If the boundary is not exceeded, we choose the updated connected component as the final moving object region.


\section{EXPERIMENTS}
\subsection{Experiment Setup}
\noindent \textbf{Datasets.} We conduct extensive experiments on the public dataset EVIMO2 \cite{mitrokhin2019ev} and self-collected various-illumination event (VIE) dataset. Different from the public dataset EVIMO2, the proposed VIE dataset contains various ego-motion patterns (\emph{i.e.}, rotation and translation) and fast moving objects under various illumination conditions (\emph{i.e.}, ideal light and low light conditions), where we utilize DAVIS346 camera to collect the paired RGB-event data with manually handcrafted moving object bounding box labels.

\noindent \textbf{Comparison Methods.} We choose the optimization-based methods (Mitrokhin \emph{et al.} \cite{mitrokhin2018event} and Zhou \emph{et al.} \cite{zhou2021event}) and the IMU-based method (Zhao \emph{et al.} \cite{zhao2023event}) for fair comparison. The optimization-based methods use the motion estimated from the events for compensation and moving object detection, while the IMU-based method uses the readout data from the hardware IMU. In addition, we choose the Intersection over Union (IoU) and the accuracy of the moving object detection as the quantitative metric.

\begin{table}
    \setlength{\abovecaptionskip}{5pt}
    \setlength\tabcolsep{4pt}
    \setlength{\belowcaptionskip}{5pt}
  \centering
    \caption{Quantitative results on different datasets.}
  \renewcommand\arraystretch{1.1}
  \begin{tabular}{cc|cccc}
    \hline
    \hline
  \multicolumn{1}{c|}{\multirow{2}{*}{Method}}&
  \multicolumn{2}{c|}{EVIMO2 \cite{mitrokhin2019ev}} &
  \multicolumn{2}{c}{VIE} \\
  \cline{2-5}
\multicolumn{1}{c|}{}& \multicolumn{1}{c|}{IoU} &
\multicolumn{1}{c|}{Accuracy} & \multicolumn{1}{c|}{IoU} &
\multicolumn{1}{c}{Accuracy} \\
\hline
\multicolumn{1}{c|}{Mitrokhin \emph{et al.} \cite{mitrokhin2018event}}& \multicolumn{1}{c|}{0.61} &
\multicolumn{1}{c|}{75.00\%} & \multicolumn{1}{c|}{ 0.54} &
\multicolumn{1}{c}{60.41\%} \\
\hline
\multicolumn{1}{c|}{Zhou \emph{et al.} \cite{zhou2021event}}& \multicolumn{1}{c|}{0.63} &
\multicolumn{1}{c|}{71.42\%} & \multicolumn{1}{c|}{0.57} &
\multicolumn{1}{c}{64.92\%} \\
\hline
\multicolumn{1}{c|}{Zhao \emph{et al.} \cite{zhao2023event}}& \multicolumn{1}{c|}{0.62} &
\multicolumn{1}{c|}{75.00\%} & \multicolumn{1}{c|}{0.62} &
\multicolumn{1}{c}{66.70\%} \\
\hline
\multicolumn{1}{c|}{Ours}& \multicolumn{1}{c|}{\textbf{0.64}} &
\multicolumn{1}{c|}{\textbf{78.57\%}} & \multicolumn{1}{c|}{\textbf{0.71}} &
\multicolumn{1}{c}{\textbf{79.95\%}} \\
\hline
\hline

  \end{tabular}
   \label{Quantitative_Comparison}
\end{table}

\subsection{Comparison Experiments}
In Fig. \ref{Comparison} and Table \ref{Quantitative_Comparison}, we show the qualitative and quantitative comparison on public and self-collected datasets. We have two conclusions.
First, our method and the IMU-based method Zhao \emph{et al.} \cite{zhao2023event} significantly outperform the optimization-based methods Mitrokhin \emph{et al.} \cite{mitrokhin2018event} and Zhou \emph{et al.} \cite{zhou2021event} by a large margin. The main reason is that, optimization methods require the ideal handcraft features, while there exist varying degrees of noise in dynamic event stream may weaken the discriminative features. On the contrary, IMU that is robust to the environment is able to provide accurate measure values for motion compensation. Second, the proposed method performs better than the IMU-based method Zhao \emph{et al.} \cite{zhao2023event}.
This is because Zhao \emph{et al.} \cite{zhao2023event} only extracts the moving object via the spatial motion confidence segmentation, limiting the structure integrity of the moving object. In contrast, the proposed method could further improve moving object detection by extracting object structure from the event point cloud.

\begin{table}
    \setlength{\abovecaptionskip}{5pt}
    \setlength\tabcolsep{4pt}
    \setlength{\belowcaptionskip}{5pt}
  \centering
    \caption{Ablation study on spatial and temporal reasoning.}
  \renewcommand\arraystretch{1.1}
  \begin{tabular}{c|c|c}
    \hline
    \hline
  Methods & IoU & Accuracy \\
  \hline
  Frame difference & 0.21 & 0\% \\
\hline
  Confidence-based spatial reasoning & 0.69 & 76.16\% \\
  \hline
  Structure-based temporal reasoning & 0.59 & 71.15\% \\
  \hline
  Joint spatio-temporal reasoning (ours) & \textbf{0.71} & \textbf{79.95\%} \\
  \hline
  \hline

  \end{tabular}
   \label{Ablation_Reasoning}
\end{table}

\subsection{Ablation Study and Discussion}
\noindent \textbf{Effectiveness of Spatio-Temporal Reasoning Framework.}
In Table \ref{Ablation_Reasoning}, we demonstrate the effectiveness of the joint spatio-temporal reasoning framework. Without any reasoning approach, we only use the frame difference method \cite{singla2014motion} to detect the moving object, which hardly performs well. With only confidence-based spatial reasoning, the object performance improves significantly. With only structure-based temporal reasoning, the object performance is slightly weaker than only spatial reasoning. The reason is that, although the object extracted from event point cloud has a complete structure, the moving object region is hollow, limiting the performance. Instead, the proposed joint spatio-temporal reasoning can fuse the advantages of the two reasoning strategies for improving the moving object detection.

\begin{figure}
  \setlength{\abovecaptionskip}{5pt}
  \setlength{\belowcaptionskip}{-5pt}
  \centering
   \includegraphics[width=0.98\linewidth]{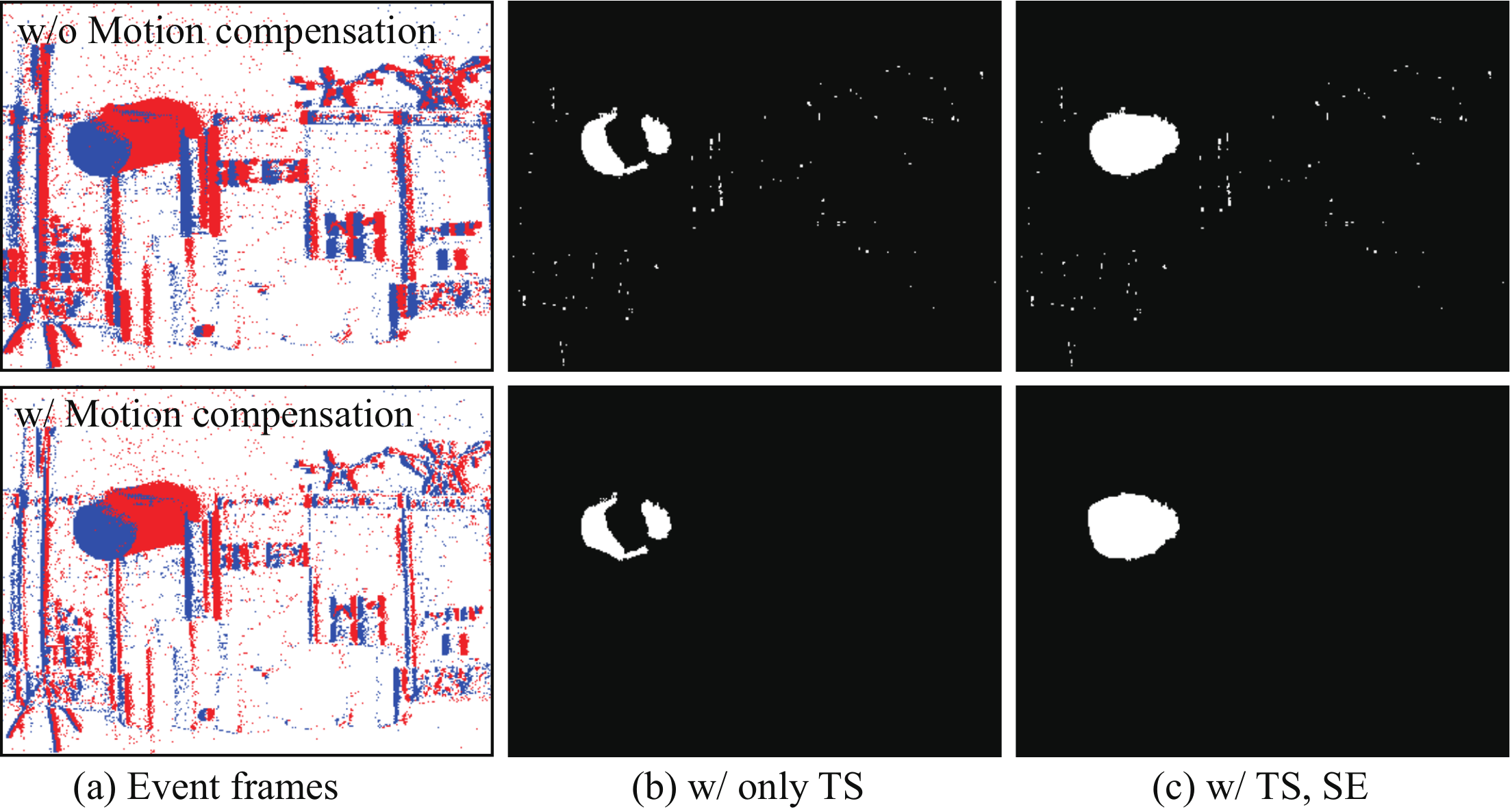}
   \caption{Effectiveness of different modules. (a) Event frames with and without motion compensation. (b) Detection results with only threshold segmentation (TS). (c) Detection results with threshold segmentation and structure extraction (SE).
   }
   \label{Ablation_Modules}
 \end{figure}

\noindent \textbf{How do Various Modules Improve Detection?}
We conduct an ablation study on the impact of different modules on the final moving object detection results in Fig. \ref{Ablation_Modules}. Motion compensation can align the background events so that background boundaries are sharp. Confidence threshold segmentation can extract the location of the moving object according to the motion confidence, while the object structure is incomplete. Structure extraction from the event point cloud could further refine the contour of the extracted object, improving moving object detection.

\noindent \textbf{Adaptability on Various Ego Motion Patterns.}
In Table \ref{Ablation_Egomotion}, we analyze the adaptability of the IMU-based method Zhao \emph{et al.} \cite{zhao2023event} and our method to different ego motion patterns of camera itself. With only the ego motion of rotation, both the Zhao \emph{et al.} \cite{zhao2023event} and our method can perform well. With only the ego motion of translation, the performance of the IMU-based method declines seriously, while the proposed method can still maintain a reliable performance. Under the complex ego motion including rotation and translation, our performance only declines slightly. This is because IMU only provides globally identical translation motion, and ego motion of translation makes it difficult to align the background events at different depths. On the contrary, the proposed method can exploit the regular structure of the moving object from the event point cloud to improve detection performance.

\begin{table}
    \setlength{\abovecaptionskip}{5pt}
    \setlength\tabcolsep{7pt}
    \setlength{\belowcaptionskip}{5pt}
  \centering
    \caption{Discussion on the impact of various ego motion patterns of the camera on moving object detection.}
  \renewcommand\arraystretch{1.1}
  \begin{tabular}{c|c|c|c}
    \hline
    \hline
  Rotation & Translation & Zhao \emph{et al.} \cite{zhao2023event} & Ours \\
  \hline
  $\times$ & $\times$ & 100\% & 100\% \\
\hline
  $\surd$ & $\times$ & 80.23\% & \textbf{87.36\%} \\
  \hline
  $\times$ & $\surd$ & 62.51\% & \textbf{81.82\%} \\
  \hline
  $\surd$ & $\surd$ & 54.65\% & \textbf{80.00\%} \\
  \hline
  \hline
  \end{tabular}
   \label{Ablation_Egomotion}
\end{table}

\begin{figure}
  \setlength{\abovecaptionskip}{5pt}
  \setlength{\belowcaptionskip}{-5pt}
  \centering
   \includegraphics[width=0.99\linewidth]{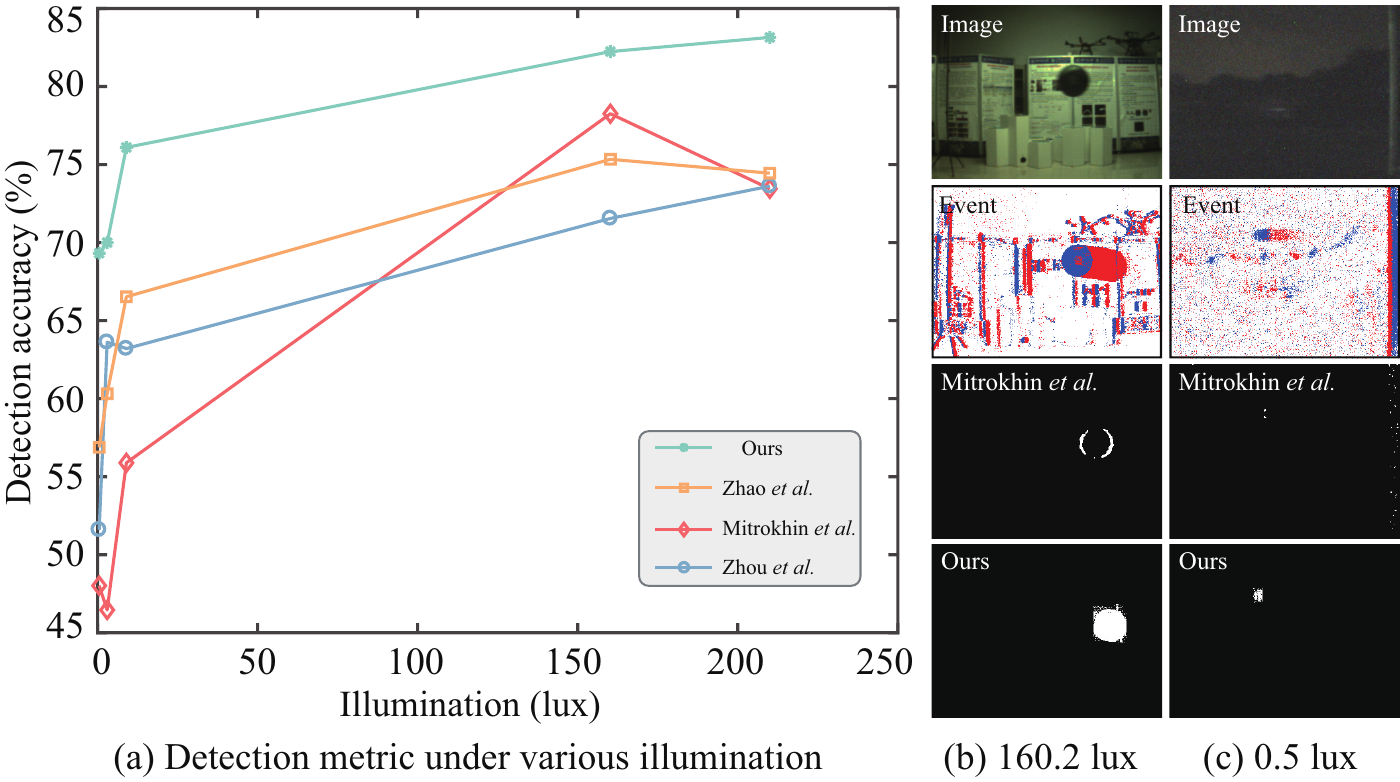}
   \caption{Generalization for various illumination conditions. (a) Comparison curve of different methods under various illumination. (b) Visualization under ideal light (160.2 lux). (c) Visualization under low light (0.5 lux).
   }
   \label{Ablation_Genelazation}
 \end{figure}

\noindent \textbf{Generalization for Different Conditions.}
In Fig. \ref{Ablation_Genelazation}, we discuss the generalization of the proposed method for different illumination conditions. In Fig. \ref{Ablation_Genelazation} (a), we set five illumination conditions, including three low light conditions and two ideal light conditions. As shown in Fig. \ref{Ablation_Genelazation} (a)-(c), we can conclude that, the optimization-based methods perform well under ideal illumination, while suffering degradation caused by the noise events under low light conditions, thus interfering with the segmentation results. On the contrary, with the help of IMU, the proposed method shows a good performance in both ideal and low light conditions.

\noindent \textbf{Limitation.}
The proposed method could extract the independent moving object with regular geometric shapes. However, in autonomous driving, though our method can effectively locate the object, it is ineffective in ensuring the structure integrity of irregular objects (\emph{e.g.}, pedestrians). The reason is that irregular objects do not follow the regular structure in the event point cloud, resulting in the ineffectiveness of the structure-based temporal reasoning. In the future, we will introduce deep models to non-linearly learn the temporal structure of moving objects to improve the robustness of the proposed method in various practical scenes.

\section{CONCLUSIONS}
In this work, we propose a novel joint spatio-temporal reasoning framework for moving object detection from motion confidence and geometry structure. To the best of our knowledge, we are the first to discover the columnar structure of the moving object in the event point cloud and investigate the RANSAC-based regular structure extraction for moving object detection. The proposed method can effectively extract the moving object with complete contour from dynamic event stream. We demonstrate that the proposed method significantly outperforms the state-of-the-art methods on both the public and self-collected datasets. We believe that our work could not only facilitate the development of moving object detection but also enlighten the researchers of the broader filed, \emph{i.e.}, scene understanding under adverse conditions.

\section*{ACKNOWLEDGMENT}
This work was supported in part by the National Natural Science Foundation of China under Grant 62371203, in part by the National Natural Science Foundation of China under Grant 62176100, and in part by Xiaomi Young Talents Program.

\addtolength{\textheight}{-9.4cm}

{
\small
\bibliographystyle{ieee_fullname}
\bibliography{egbib}

}

\end{document}